\documentclass{article} 
\usepackage{iclr2017_workshop,times}
\usepackage{hyperref}
\usepackage{url}
\usepackage{amsmath}
\usepackage{graphicx}

\title{Factorization tricks for LSTM networks}

\author{Oleksii Kuchaiev \\
NVIDIA \\
\texttt{okuchaiev@nvidia.com} \\ 
\And
Boris Ginsburg \\
NVIDIA\\
\texttt{bginsburg@nvidia.com} \\
}

\begin{document}

\maketitle

\begin{abstract}
We present two simple ways of reducing the number of parameters and accelerating the training of large Long Short-Term Memory (LSTM) networks: the first one is "matrix factorization by design" of LSTM matrix into the product of two smaller matrices, and the second one is partitioning of LSTM matrix, its inputs and states into the independent groups. Both approaches allow us to train large LSTM networks significantly faster to the near state-of the art perplexity while using significantly less RNN parameters.
\end{abstract}

\section{Introduction}
LSTM networks  \citep{hochreiter1997long} have been successfully used in language modeling  \citep{jozefowicz2016exploring,shazeer2017outrageously}, speech recognition \citep{xiong2016achieving}, machine translation \citep{wu2016google}, and many other tasks. However, these networks have millions of parameters, and require weeks of training on multi-GPU systems.

We introduce two modifications of LSTM cell with projection, LSTMP \citep{sak2014long}, to reduce the number of parameters and speed-up training. The first method, \textit{factorized LSTM} (F-LSTM) approximates big LSTM matrix with a product of two smaller matrices. The second method, \textit{group LSTM} (G-LSTM) partitions LSTM cell into the independent groups. We test  F-LSTM and G-LSTM architectures on the task of language modeling using One Billion Word Benchmark \citep{chelba2013one}. As a baseline, we used BIGLSTM model without CNN inputs described by \citet{jozefowicz2016exploring}. We train all networks for 1 week on a DGX Station system with 4 Tesla V100 GPUs, after which BIGLSTM's evaluation perplexity was 35.1. Our G-LSTM based model got 36 and F-LSTM based model got 36.3 while using two to three times less RNN parameters.

\subsection{Long Short-Term Memory overview}
Learning long-range dependencies with Recurrent Neural Networks (RNN) is challenging due to the vanishing and exploding gradient problems \citep{bengio1994learning, pascanu2013difficulty}. To address this issue, the  LSTM cell has been introduced by  \citet{hochreiter1997long}, with the following recurrent computations:  
\begin{equation}\label{eq:1}
LSTM:  h_{t-1}, c_{t-1}, x_{t}  \rightarrow h_t, c_t.
\end{equation}
where $x_t$ is input, $h_t$ is cell's state, and $c_t$ is cell's memory. We consider LSTM cell with projection of size $p$, LSTMP, where Equation \ref{eq:1} is computed as follows \citep{sak2014long, zaremba2014recurrent}.
 First, cell gates $(i,f,o,g)$ are computed:
\begin{equation} \label{eq:2}
\begin{pmatrix} i \\ f \\ o \\ g  \end{pmatrix} = \begin{pmatrix} sigm \\ sigm \\ sigm \\ tanh  \end{pmatrix} T \begin{pmatrix} x_t \\ h_{t-1} \end{pmatrix} 
\end{equation}
where $x_t \in R^p$, $h_t \in R^p$, and $T:R^{2p} \rightarrow R^{4n}$ is an affine transform $T=W*[x_{t}, h_{t-1}] + b$. 

Next state $h_t \in R^p$ and memory $c_t \in R^n$ are computed using following equations:
$$
c_t = f\odot c_{t-1} + i\odot g ; \ \ \ 
h_t = P (o\odot tanh(c_t))
$$
where $P: R^n \rightarrow R^p$ is a linear projection. The major part of LSTMP cell computation is in computing affine transform $T$ because it involves multiplication with $4n\times 2p$ matrix $W$. Thus we focus on reducing the number of parameters in $W$.

\subsection{Related Work}
The partition of layer into parallel groups have been introduced by \citet{krizhevsky2012imagenet} in AlexNet, where some convolutional layers have been divided into two groups to split the model between two GPUs.  Multi-group convnets have been widely used to reduce network weights and required compute, for example by \citet{esser2016convolutional}. This multi-group approach was extended to the extreme in  Xception architecture by \citet{chollet2016xception}.
The idea of factorization of large convolutinal layer into the stack of layers with smaller filters was used, for example, in VGG networks \citep{simonyan2014very}, and in ResNet ``bottleneck design'' \citep{He_2016_CVPR}. \citet{denil2013predicting} have shown that it is possible to train several different deep architectures by learning only a small number of weights and predicting the rest.
In case of LSTM networks, ConvLSTM \citep{Shi:2015:CLN:2969239.2969329}, has been introduced to better exploit possible spatiotemporal correlations, which is conceptually similar to grouping.

\section{Models}

\subsection{Factorized LSTM cell}
Factorized LSTM (F-LSTM) replaces matrix $W$  by the product of two smaller matrices that essentially try to approximate $W$ as $W\approx W2*W1$, where $W1$ is of size $ 2p \times r$, $W2$ is  $r \times 4n$, and  $r<p<=n$ ("factorization by design"). The key assumption here is that $W$ can be well approximated by the matrix of rank $r$. Such approximation contains less LSTMP parameters than original model - $(r*2p+r*4n)$ versus $(2p*4n)$ and, therefore, can be computed faster and synchronized faster in the case of distributed training.
\begin{figure}[h]
\begin{center}
\includegraphics[height=4cm]{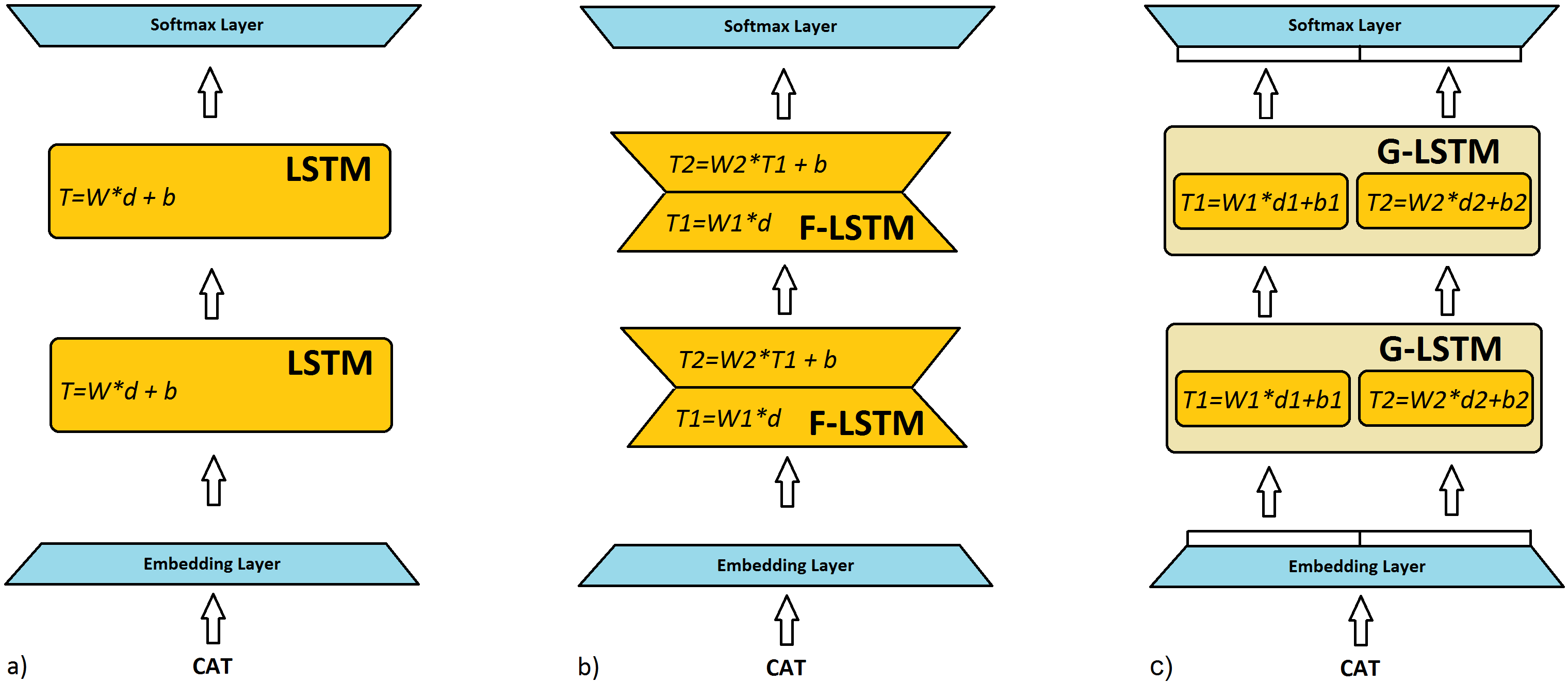}
\end{center}
\caption{ Language model using: (a) 2 regular LSTM layers, (b) 2 F-LSTM layers, and (c) 2 G-LSTM layers with 2 group in each layer. Equations inside cells show what kind of affine transforms are computed by those cells at each time step. Here $d=(x,h)$ for models without groups and $d1=(x^1,h^1)$, $d2=(x^2,h^2)$ for model with two groups; and time index dropped for clarity.}
\label{fig:glstm}
\end{figure}
\subsection{Group LSTM cell}
This approach is inspired by groups in Alexnet \citep{krizhevsky2012imagenet}. We postulate that some parts of the input $x_t$ and hidden state $h_t$ can be thought of as independent feature groups. For example, if we use two groups, then both $x_t$ and $h_t$ are effectively split into two vectors concatenated together $x_t=(x^1_t, x^2_t)$ and $h_t=(h^1_t, h^2_t)$, with $h^i_t$ only dependent on $x^i_t$, $h^i_{t-1}$ and cell's memory state. Therefore, for $k$ groups Equation \ref{eq:2} changes to:
\begin{equation}
\begin{pmatrix} i \\ f \\ o \\ g  \end{pmatrix} = \begin{pmatrix} \begin{pmatrix} sigm \\ sigm \\ sigm \\ tanh  \end{pmatrix} T^1 \begin{pmatrix} x^1_t \\ h^1_{t-1} \end{pmatrix}, ... , \begin{pmatrix} sigm \\ sigm \\ sigm \\ tanh  \end{pmatrix} T^k\begin{pmatrix} x^k_t \\ h^k_{t-1} \end{pmatrix} \end{pmatrix}
\end{equation}
where, $T^j$ is a group $j$'s affine transform from $R^{2p/k}$ to $R^{4n/k}$. The partitioned $T$ will now have $k*\frac{4n*2p}{k*k}$ parameters. This cell architecture is well suited for model parallelism since every group computation is independent.
An alternative interpretation of G-LSTM layers is demonstrated in the Figure \ref{fig:glstm} (c). 
While this might look similar to ensemble \citep{shazeer2017outrageously} or multi-tower \citep{ciregan2012multi} models, the key differences are: (1) input to different groups is different and assumed independent, and (2) instead of computing ensemble output, it is concatenated into independent pieces. 

\section{Experiments and Results} \label{expsection}

For testing we used the task of learning the joint probabilities over word sequences of arbitrary lengths $n$: 
$
P(w_1,...,w_n)=\prod_{i=1}^{n}P(w_i|w_1,...,w_{i-1})
$, such that ``real'' sentences have high probabilities compared to the random sequences of words. Figure \ref{fig:glstm} (a) shows the typical LSTM-based model, where first the words are embedded into the low dimensional dense input for RNN, then the ``context'' is learned using RNNs via number of steps and, finally, the softmax layer converts RNN output into the probability distribution $P(w_1,...,w_n)$.
We test the following models: 
\begin{itemize} 
\item BIGLSTM -  model with projections but without CNN inputs from \citet{jozefowicz2016exploring}
\item BIG F-LSTM F512 - with intermediate rank of 512 for LSTM matrix $W$, \item BIG G-LSTM G-4, with 4 groups in both layers 
\item BIG G-LSTM G-16,  with 16 groups in both layers. 
\end{itemize} 

We train all models on DGX Station with 4 GV100 GPUs for one ween using Adagrad optimizer, projection size of 1024, cell size of 8192, mini-batch of 256 per GPU, sampled softmax with 8192 samples and 0.2 learning rate. Note that the use of projection is crucial as it helps to keep down embedding and softmax layer sizes. Table \ref{1bwbm-1week} summarizes our experiments.

Judging from the training loss Plots  \ref{fig:exp1plot} in \textit{Appendix}, it is clearly visible that at the same step count, model with more parameters wins. However, given the same amount of time, factorized models train faster. While the difference between BIGLSTM and BIG G-LSTM-G2 is clearly visible, BIG G-LSTM-G2 contains almost 2 times less RNN parameters than BIGLSTM, trains faster and, as a results, achieves similar evaluation perplexity within the same training time budget (1 week).

Our code is available at \url{https://github.com/okuchaiev/f-lm} 

\begin{table}[t]
\caption{One Billion Words benchmark evaluation results after 1 week of training using one DGX Station with 4 Tesla V100 GPUs.}
\label{1bwbm-1week}
\begin{center}
\begin{tabular}{lllll}
\multicolumn{1}{c}{\bf Model}  &\multicolumn{1}{c}{\bf Perplexity}  &\multicolumn{1}{c}{\bf Step} &\multicolumn{1}{c}{\bf Num of RNN parameters} &\multicolumn{1}{c}{\bf Words/sec}
\\ \hline \\
BIGLSTM baseline  & 35.1         & 0.99M & 151,060,480 & 33.8K\\
BIG F-LSTM F512   & 36.3         & 1.67 M & 52,494,336 & ~56.5K\\
BIG G-LSTM G-2    & 36           & 1.37M  & 83,951,616 & ~41.7K\\
BIG G-LSTM G-4    & 40.6         & 1.128M & 50,397,184 & ~56K\\
BIG G-LSTM G-8    & 39.4         & 850.4K & 33,619,968 & ~58.5K\\
\end{tabular}
\end{center}
\end{table}
\subsection{Future research}

While one might go further and try to approximate  transform $T$ using arbitrary feed forward neural network with $2p$ inputs and $4n$ outputs, during our initial experiments we did not see immediate benefits of doing so. Hence, it remains a topic of future research.

It might be possible to reduce the number of RNN parameters even further by stacking G-LSTM layers with increasing group counts on top of each other. 
In our second, smaller experiment, we replace the second layer of BIG G-LSTM-G4 network by the layer with 8 groups instead of 4, and call it BIG G-LSTM-G4-G8. We let both BIG G-LSTM-G4 and BIG G-LSTM-G4-G8 ran for 1 week on 4 GPUs each and achieved very similar perplexities. Hence, the model with ``hierarchical'' groups did not lose much accuracy, ran faster and got better perplexity. Such ``hierarchical'' group layers look intriguing as they might provide a way for learning different levels of abstractions but this remains a topic of future research.

{\bf Acknowledgements} 
We are grateful to Scott Gray and Ciprian Chelba for helping us identify and correct issues with earlier versions of this work.

\bibliography{citations}
\bibliographystyle{iclr2017_workshop}

\pagebreak
\section*{Appendix: Training loss for 4 LSTM-like models} \label{appendix}

\begin{figure}[h]
\begin{center}
\includegraphics[height=10cm]{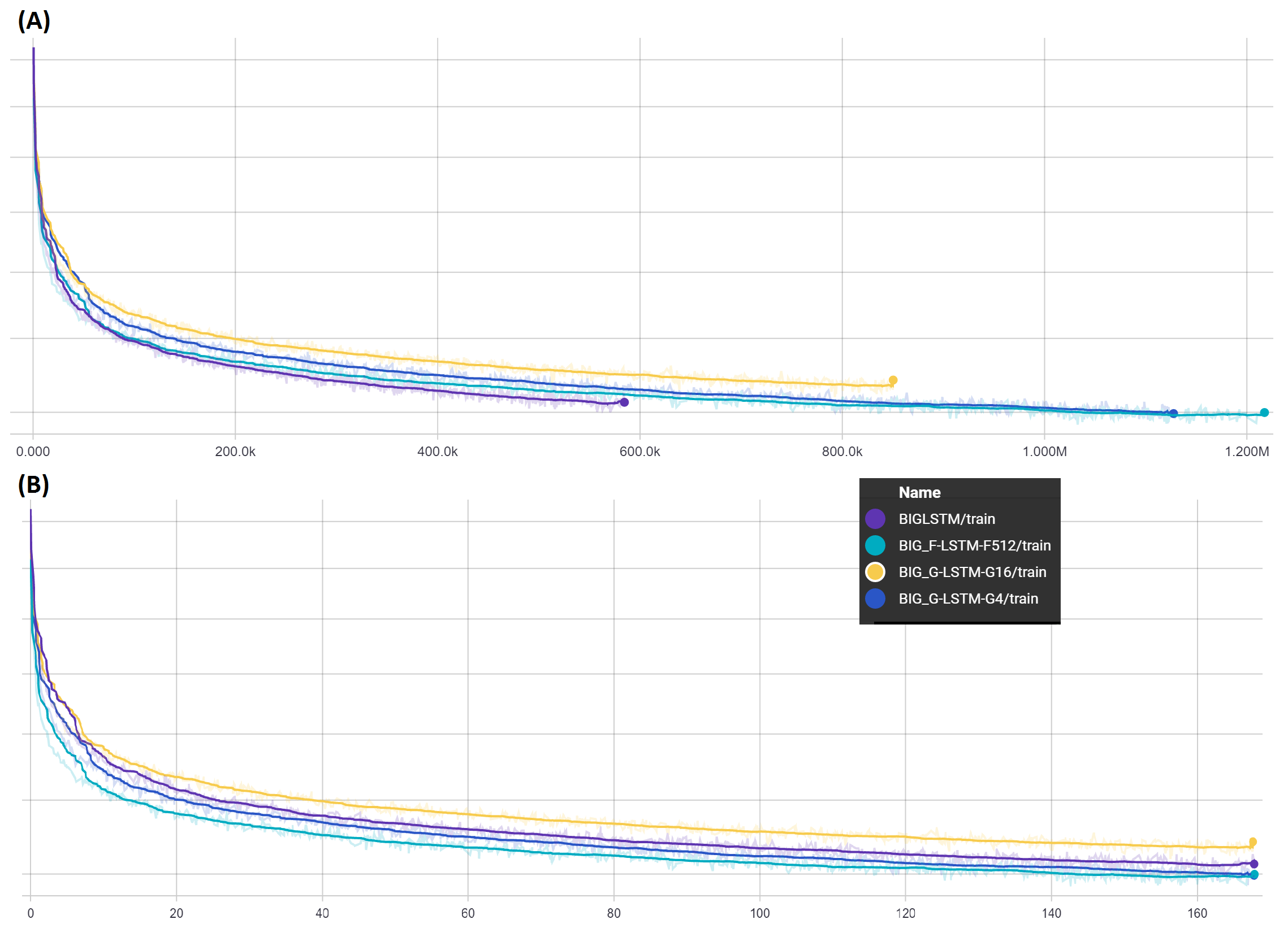}
\end{center}
\caption{Y-axis: same for (A) and (B) - training loss log-scale, X-axis: for (A) - step, or mini-batch count, for (B) - hours (w.g. wall time) of training. BIGLSTM baseline, BIG G-LSTM-G4, BIG G-LSTM-G16, and BIG F-LSTM-F512 all trained for exactly one week. It is clearly visible, that at the same step count, the model with more parameters wins. On the other hand, factorized models can do significantly more iterations in the given amount of time and therefore get to the better results given same amount of time. (full extent of X-axis for both (A) and (B) is 1 week).}
\label{fig:exp1plot}
\end{figure}

\end{document}